\documentclass{article}

\usepackage[preprint]{neurips_2026}
\usepackage{enumitem}
\usepackage[utf8]{inputenc} 
\usepackage[T1]{fontenc}    
\usepackage{hyperref}       
\usepackage{url}            
\usepackage{booktabs}       
\usepackage{amsfonts}       
\usepackage{nicefrac}       
\usepackage{microtype}      
\usepackage{xcolor}         
\usepackage{graphicx}       
\usepackage{amsmath,amssymb}

\title{Sensoformer: Robust Sim-to-Real Inference on Variable-Geometry Sensor Sets via Physics-Structured Randomization}

\author{%
  Zhe Jia\thanks{Corresponding author.} \\
  Institute for Geophysics\\
  University of Texas at Austin\\
  Austin, USA \\
  \texttt{zjia@ig.utexas.edu} \\
  \And
  Jim Zhang (Xiaotian) \\
  Institute for Geophysics\\
  University of Texas at Austin\\
  Austin, USA \\
  \texttt{jxzhang1998@gmail.com} 
  \And
  Junpeng Li \\
  Institute for Geophysics \& \\
  Dept. of Earth and Planetary Sciences\\
  University of Texas at Austin\\
  Austin, USA \\
  \texttt{jl88749@my.utexas.edu} \\ 
}

\begin{document}

\maketitle

\begin{abstract}
Inferring high-dimensional physical states from sparse, ad-hoc sensor arrays is a fundamental challenge across AI for Science and industrial IoT. Standard machine learning architectures struggle in these domains due to irregular, variable-cardinality sensor geometries and the profound sim-to-real distribution shift caused by unmodeled physical heterogeneities. To address these challenges, we propose Sensoformer, a set-attention framework integrated with Physics-Structured Domain Randomization (PSDR). By explicitly randomizing the underlying physical dynamics (e.g., propagation media, extreme noise, and network availability dropout) rather than just visual features, PSDR enforces the learning of domain-invariant physical operators. Using seismic source inversion as a rigorous real-world testbed, Sensoformer is pre-trained on 100,000 synthetics and evaluated on a highly complex real-world catalog. We demonstrate that Sensoformer achieves state-of-the-art precision and outperforms Message Passing Neural Networks (MPNNs) and Neural Operators (e.g., DeepONet) which struggle with extreme spatial sparsity and mixed-modality inputs. Furthermore, interpretability analysis reveals that the attention mechanism autonomously discovers optimal experimental design principles, dynamically prioritizing sparse orthogonal sensors to overcome information bottlenecks.
\end{abstract}

\section{Introduction}
How do we reconstruct a complex physical event, whether it is a distant earthquake, a cosmic radio burst, or a climate anomaly, using only a handful of scattered sensors?\footnote{Code and data are available at: \url{https://github.com/jiazhe868/Sensoformer/}} This is the archetypal challenge across the physical sciences and industrial Internet of Things (IoT) \citep{bergen2019machine,bouman2016computational,ravuri2021skilful}: inferring complex physical source parameters $\theta$ from a sparse set of indirect, noisy observations $X = \{x_1, \dots, x_n\}$. Unlike images or video where data pixels are fixed on a regular grid, scientific data is often collected by ad-hoc sensor arrays where the number of sensors varies per event and their spatial arrangement is irregular.

While classical inverse solvers \citep{tarantola2005inverse} rely on iterative optimization and are computationally costly for real-time applications, deep learning offers a compelling alternative via amortized inference \citep{cranmer2020frontier}. However, there is a fundamental geometric mismatch between standard AI architectures and physical sensor data. Convolutional Neural Networks (CNNs) \citep{lecun2015deep} require rigid grids that force artificial interpolation. Graph Neural Networks (GNNs/MPNNs) can handle irregularity but typically rely on local message-passing, which struggles to capture global physical constraints efficiently. Furthermore, emerging Neural Operators (e.g., DeepONet) map continuous function spaces but often assume dense query points, struggling with extreme, dynamic spatial sparsity and mixed-modality discrete metadata. To address these challenges, we need highly flexible, permutation-invariant set architectures capable of modeling global pairwise relationships \citep{zaheer2017deep, lee2019set}.

Compounding the geometric challenge is the Sim-to-Real gap \citep{tobin2017domain}. In science disciplines like geophysics and robotics, obtaining ground-truth labels for real-world observations is expensive and labor-intensive, therefore models have to be trained on physics-based simulations, which inevitably lack the complex realities of unmodeled propagation effects and structural heterogeneity \citep{karniadakis2021physics}. Naive training on idealized simulations faces a substantial out-of-distribution (OOD) shift. A trustworthy AI for Science framework must learn invariant physical operations that hold true despite these distribution shifts.

Here, we present Sensoformer, an end-to-end, interpretable framework for solving inverse problems on variable sensor sets. We utilize seismic source inversion as a rigorous, highly complex real-world benchmark to validate our methodology. We have three methodological advancements:
\begin{enumerate}[leftmargin=*, itemsep=1pt]
\item We bridge the Sim-to-Real gap not by feature alignment, but by randomizing the physics of the generating process (e.g., velocity models, sensor dropouts), forcing the model to learn representations invariant to unmodeled structural heterogeneity. We validate that the gap is bridged through manifold alignment.
\item We demonstrate that global self-attention effectively resolves spatial ambiguities inherent in wave physics. With comprehensive benchmarking, Sensoformer outperforms pooling-based architectures (DeepSets), local message-passing networks (MPNNs), and neural operators (DeepONet), achieving state-of-the-art precision under extreme spatial sparsity.
\item Through Explainable AI (XAI) analysis, we reveal that the learned attention mechanism acts as a dynamic information filter. It autonomously identifies the geometric information bottleneck of the sensor network and learns a global policy that prioritizes sparsely distributed orthogonal sensors, effectively recovering principles of optimal experimental design purely from data.
\end{enumerate}

\section{Related Work}
\subsection{Deep Learning on Sets and Irregular Grids}
Handling variable-cardinality inputs requires permutation-invariant architectures like DeepSets \citep{zaheer2017deep} and PointNet \citep{qi2017pointnet}. However, their reliance on global pooling (e.g., $\mathrm{sum}$ or $\mathrm{max}$) is often too lossy for complex physical systems where relative spatial dependencies and pairwise contexts are critical. Graph Neural Networks (GNNs) \citep{kipf2016semi} model dependencies explicitly but rely on local message-passing, which can over-smooth features and struggle to capture global constraints efficiently. Neural Operators (e.g., DeepONet \citep{lu2021learning}, FNO \citep{li2020fourier}) map infinite-dimensional spaces but typically assume dense sampling on fixed grids, struggling with extreme, dynamic spatial sparsity and mixed-modal inputs. In contrast, Set Transformers \citep{lee2019set} leverage global self-attention. \textbf{Sensoformer} adapts this by modeling all-to-all pairwise interactions dynamically, preserving essential geometric constraints without imposing rigid topologies.

\subsection{Sim-to-Real Transfer and Domain Randomization}
Bridging the sim-to-real gap is a fundamental challenge across robotics and the physical sciences \citep{tobin2017domain, karniadakis2021physics}. While unsupervised domain adaptation aligns feature distributions (e.g., via adversarial training \citep{ganin2016domain}), it requires target domain access during training and can be unstable. Domain Randomization (DR) \citep{tobin2017domain} instead forces models to learn invariants by randomizing simulator parameters. While standard DR focuses on visual or contact dynamics \citep{akkaya2019solving}, scientific inference faces uncertainties in the governing physics themselves. Sensoformer extends DR into Physics-Structured Domain Randomization (PSDR) by randomizing the core generative processes (e.g., structural heterogeneity, complex scattering). This provides a generalizable blueprint for learning robust physical operators despite highly imperfect simulations.

\subsection{Machine Learning for Physical Inverse Problems}
While deep learning has transformed scientific discriminative tasks (e.g., phase picking in seismology \citep{zhu2019phasenet}), end-to-end parameter inversion from sparse arrays remains an open challenge. Standard CNN-based solvers \citep{kuang2021real} impose artificial grids, limiting their generalization to ad-hoc real-world networks. Recent approaches employing GNNs \citep{zhang2022spatiotemporal} address geometric irregularity but often treat sensors as independent evidence, ignoring the long-range directional dependencies between distant sensors. Meanwhile, early Transformer applications in this domain \citep{munchmeyer2021transformer} treat sensors as tokens yet rely on discrete positional encodings, lacking the continuous spatial resolution required for precise physical inference. Sensoformer overcomes these limitations by utilizing continuous physical metadata to construct scale-invariant set embeddings.

\subsection{Interpretability and Optimal Experimental Design}
Building trust in scientific AI requires moving beyond standard attribution maps (e.g., Grad-CAM \citep{selvaraju2017grad}) to understand why specific data points matter \citep{rudin2019stop}. Optimal Experimental Design (OED) \citep{krause2008near} formally addresses this by organizing observations to maximize information gain and reduce uncertainty. While recent methods approximate OED via separate variational optimization loops \citep{foster2019variational}, we demonstrate that such sensing strategies can emerge implicitly within the network architecture. By analyzing our model's attention weights, we show that Sensoformer learns to dynamically prioritize sensors based on their geometric information value, effectively recovering OED principles purely from data without explicit supervision.

\section{Problem Formulation and Methodology}
We formalize the inference of complex physical states from ad-hoc sensor networks as a regression task on variable-cardinality sets. We use earthquake source inversion as our primary benchmarking task to validate the proposed framework.

\begin{figure}[t]
    \centering
    \includegraphics[width=\linewidth]{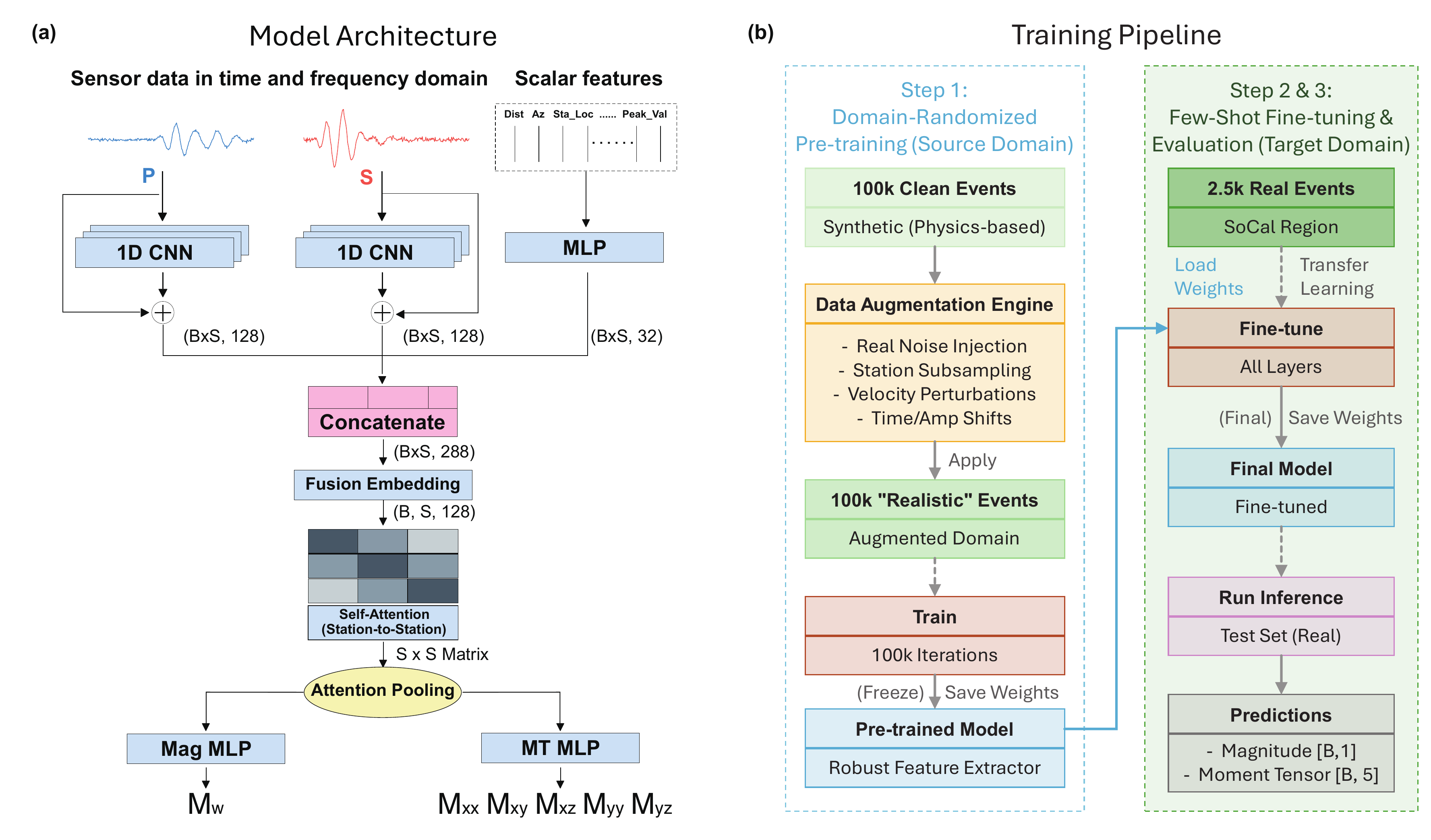} 
    \vspace{-0.15in}
    \caption{\textbf{The Sensoformer Architecture and Pipeline.} \textbf{(a)} A hierarchical Set Transformer designed for irregular sensor arrays. The model operates in three stages: (1) Multi-modal Station Encoders process dual-domain waveforms and scalar metadata into physics-aware embeddings. (2) Self-Attention models global pairwise interactions to resolve spatial ambiguities. (3) Attention Pooling dynamically weights stations based on information gain to regress the physical state. \textbf{(b)} Physics-Structured Training Curriculum. We bridge the reality gap via a two-stage protocol. First, we pre-train on synthetics augmented with aleatoric physical uncertainties (PSDR) to enforce invariant feature learning. We then fine-tune on a smaller set of real events using a weighted random sampler to correct for label imbalances. }
    \label{fig:arch}
    \vspace{-0.2in}
\end{figure}

\textbf{The Physical Target ($\mathbf{y}$)}. 
For our seismic benchmark, the target event is parameterized by a latent physical state $\mathbf{y} \in \mathbb{R}^6$. This state encapsulates the 3D Moment Tensor \citep{aki2002quantitative}, representing the fault orientation and slip direction (5 independent zero-trace $M_{ij}$ components), and the moment magnitude $M_W$ indicating the energy scale. 

\textbf{The Observation Set ($\mathbf{X}$)}. 
Unlike images defined on regular grids, real-world physical monitoring relies on ad-hoc collections of sensors. We represent the observation as an unordered set $\mathbf{X} = \{ (\mathbf{w}_i, \mathbf{s}_i) \}_{i=1}^{N}$, where $N$ is the varying number of available sensors per event. This highly variable cardinality renders standard fixed-length representations inapplicable, necessitating a permutation-invariant architecture. Each set element is a multi-modal tuple: $\mathbf{w}_i \in \mathbb{R}^{C \times T}$ contains dual-domain waveform features (raw time-series and spectral representations), while $\mathbf{s}_i \in \mathbb{R}^d$ contains discrete scalar metadata (e.g., spatial coordinates, azimuth, and distance-based amplitude decay).

\textbf{Learning Objectives}. 
We aim to learn an inverse operator $f_\theta: \mathbf{X} \to \mathbf{y}$. This function must satisfy permutation invariance, i.e., $f_\theta(\pi(\mathbf{X})) = f_\theta(\mathbf{X})$ for any permutation $\pi$ \citep{zaheer2017deep}. Due to the scarcity of high-quality real-world labels, the model is primarily trained on a simulated domain $\mathcal{D}_{\text{Sim}}$ but optimized for a target real-world domain $\mathcal{D}_{\text{Real}}$. Our strategy is to inject structured physical priors into $\mathcal{D}_{\text{Sim}}$ to learn domain-invariant operators.

\subsection{Physics-Structured Domain Randomization (PSDR)}
Bridging the sim-to-real gap in physical sciences requires more than standard data augmentation (e.g., Gaussian noise). Real-world sensor data is heavily distorted by unmodeled 3D environmental heterogeneities and complex signal scattering. To ensure robust generalization, we propose Physics-Structured Domain Randomization (PSDR).

Rather than relying on unstable domain adaptation algorithms to align features, PSDR explicitly randomizes the underlying physical generative process. We define our data generation pipeline as:
\begin{equation}
\mathbf{X}_{\text{aug}} = \mathcal{M}\left( \mathcal{T}(S(\mathbf{y}, \phi)) + \mathbf{n} \right)
\end{equation}
where $S$ is the physics-based simulator. We introduce four specific types of randomization to mimic real-world entropy (detailed hyperparameters are provided in Appendix A):

\begin{enumerate}[leftmargin=*, itemsep=1pt]
\item \textbf{Generative Environment ($\phi$):} Standard simulations assume a single, perfect medium. To prevent memorization of a specific environment, we uniformly sample from a library of 17 distinct 1D velocity models during generation. This forces the network to learn features invariant to the propagation medium.
\item \textbf{Signal Distortion ($\mathcal{T}$):} Real signals deviate from theoretical predictions due to 3D scattering. We apply stochastic phase shifts and amplitude scaling, and superimpose a simulated scattering coda (exponentially decaying tails). This regularizes the model against over-relying on idealized Green's functions.
\item \textbf{Realistic Noise Injection ($\mathbf{n}$):} Gaussian noise fails to capture complex environmental vibrations. We strictly sample noise vectors $\mathbf{n}$ from a massive library of real-world ambient recordings, teaching the model to disentangle true signals from structured background noise.
\item \textbf{Variable-Geometry Masking ($\mathcal{M}$):} To handle dynamic array geometries and telemetry failures, we apply a stochastic masking operator (sensor dropout) during training. This forces the model to recover the physical state from subsets of varying cardinality ($N \sim 30-50$).
\end{enumerate}

\begin{figure}[t!]
    \centering
    \includegraphics[width=\linewidth]{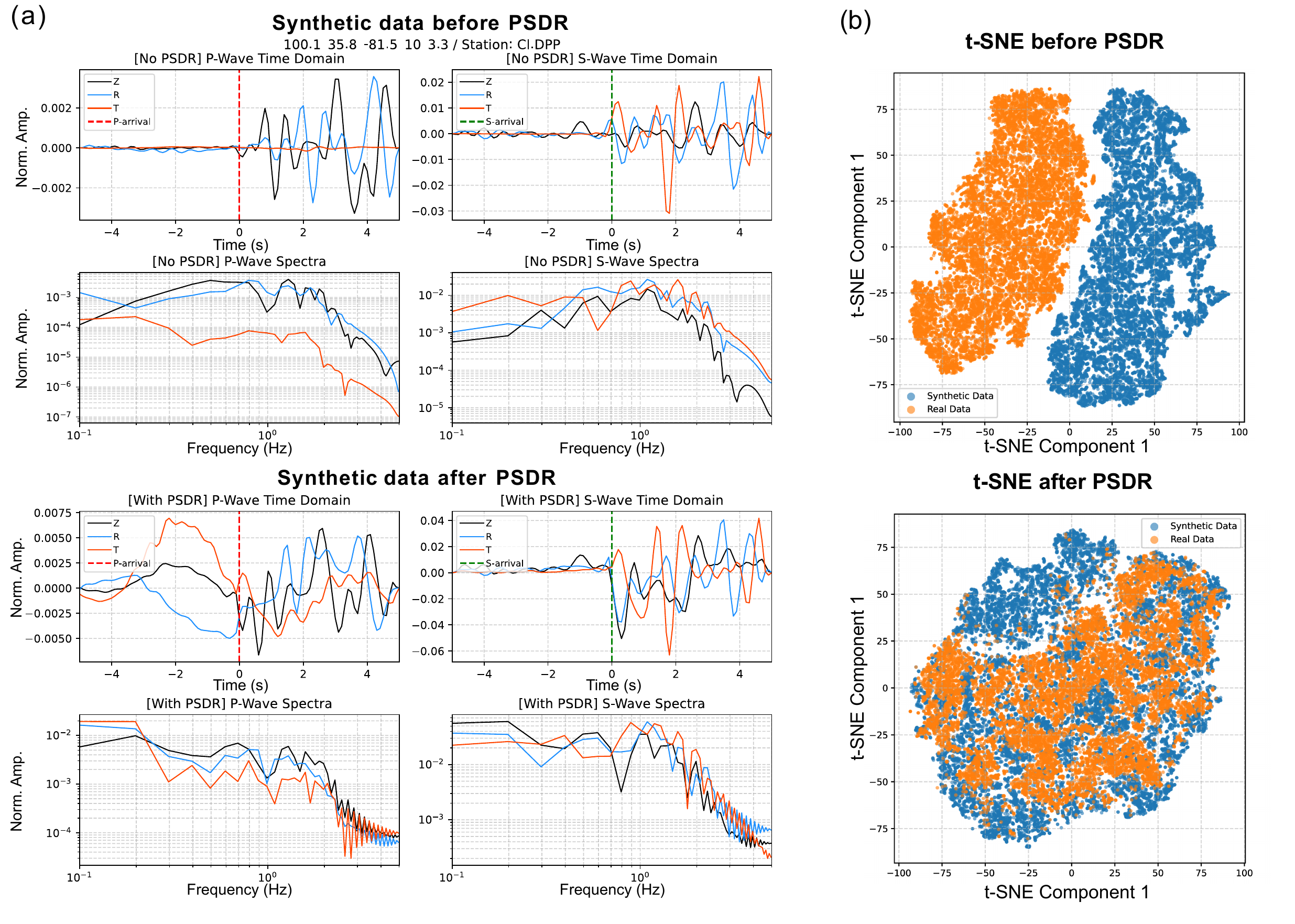}
    \vspace{-0.15in}
    \caption{\textbf{Validation of Physics-Structured Domain Randomization.} \textbf{(a)} Comparison of clean synthetic data (top) versus PSDR-augmented data (bottom). \textbf{(b)} t-SNE visualization of the feature space. A naive baseline (Left) shows disjoint clusters for synthetic and real data. In contrast, Sensoformer (Right) achieves excellent manifold alignment, demonstrating the learning of invariant physical operators robust to profound distribution shifts.}
    \label{fig:manifold}
    \vspace{-0.2in}
\end{figure}

To validate our PSDR strategy, we compare the latent spaces of a naive baseline trained on clean synthetics versus Sensoformer trained with PSDR (both fine-tuned on the same real-world subset). As shown in the t-SNE \citep{maaten2008visualizing} visualizations (Figure \ref{fig:manifold}), standard fine-tuning is insufficient to bridge the reality gap, resulting in disjoint clusters. In contrast, Sensoformer achieves perfect manifold alignment, proving that PSDR successfully extracts domain-invariant physical representations.

\subsection{Sensoformer Architecture}
Sensoformer is designed to process unstructured sensor sets through local multi-modal feature extraction followed by global relational reasoning. As illustrated in Figure \ref{fig:arch}, the architecture consists of three core modules:

\textbf{1. Multi-Modal Station Encoders (Local Feature Extraction).}
Generic time-series models treat all input channels identically. However, physical signals often consist of distinct, orthogonal components. We therefore decompose the input for the $i$-th station into three parallel streams to construct a physics-aware representation. We process the compressional (P-wave) and shear (S-wave) time-series through two parallel 1D-ResNets \citep{he2016deep}. Because these waves propagate differently and carry orthogonal physical constraints, isolating them explicitly prevents feature confounding—a necessity thoroughly validated in our ablation studies. Both time-domain and spectral-domain representations are utilized to capture high-frequency onset polarities and low-frequency coda properties. Meanwhile, standard Transformers \citep{vaswani2017attention} rely on discrete positional encodings, which fail for irregular and dynamic sensor topologies. To overcome this, we project a 20-dimensional metadata vector via an MLP. This vector includes exact spatial coordinates (lat/lon, azimuth, distance) and physical attenuation features (max amplitudes and P/S ratios). This acts as a continuous spatial and energy encoding, enabling the scale-invariant Transformer to anchor local waveforms into a global geometric context. These three views are concatenated to form a rich, local station embedding $\mathbf{h}_i$.

\textbf{2. Global Relational Reasoning via Self-Attention.}
A fundamental limitation of standard set architectures (e.g., DeepSets) is their reliance on global pooling over independently processed elements. This inherently masks pairwise relational context. In wave physics, local observations are ambiguous: a single sensor cannot resolve a global radiation pattern. We employ a Set Transformer Encoder \citep{lee2019set} to model all-to-all dependencies dynamically. Through self-attention, each station queries the entire array, contextualizing its local features against distant sensors (e.g., implicitly computing relative amplitude decays or phase differences across the spatial network). 

\textbf{3. Dynamic Information Bottleneck (Attention Pooling).}
Real-world inference on ad-hoc networks requires extreme adaptability, as the "information value" of a sensor varies drastically depending on the event's location and fault geometry. Instead of static mean or max pooling, we utilize attention pooling \citep{ilse2018attention} to compute a weighted, event-specific global representation. This allows the model to dynamically prioritize highly informative sensors and suppress redundant or noisy ones, effectively acting as an implicit spatial filter before regressing the final physical parameters.

\section{Experimental Setup}
We design our experiments to rigorously test Sensoformer's ability to learn from physically diverse simulations and transfer to a heavily biased, noisy real-world distribution. We use the Southern California seismic network as our primary real-world testbed \citep{yang2012computing}.

\textbf{Synthetic Pre-training Domain.} We generate a large-scale dataset of 100,000 synthetic events using PSDR. To ensure the model learns generalized physics, we sample physical source parameters (parameterized by the 6-component continuous moment tensor $M_{ij}$) uniformly across the full theoretical space. To mimic variable network density, for events with more than 50 available stations, we randomly sub-sample a set of $30$--$50$ stations per training instance.

\textbf{Real-world Fine-tuning Domain.} We use a high-quality catalog of 2,435 earthquakes ($M>3.0$) in Southern California \citep{yang2012computing}. These events feature real-world instrumental noise, 3D scattering effects, and highly irregular station coverage.

Our training follows a two-stage curriculum. First, Sensoformer is pre-trained for 150 epochs on the synthetic dataset (AdamW optimizer, learning rate $2 \times 10^{-4}$). Then, we fine-tune the network on the real-world dataset. To prevent catastrophic forgetting of rare physical mechanisms, we employ a weighted random sampler that balances the exposure across all target faulting types. Fine-tuning uses a lower learning rate ($2 \times 10^{-6}$).

\section{Experimental Results and Analysis}

We evaluate Sensoformer on the real-world operational catalog to measure the quality of sim-to-real transfer and architecture robustness.

\subsection{Benchmarking against ML Baselines and Domain SOTA}
To validate our architectural design, we benchmark Sensoformer against several machine learning baselines adapted for this variable-geometry task, including DeepSets \citep{zaheer2017deep}, MPNN (Message Passing Graph-based Neural Network), and DeepONet \citep{lu2021learning}. As shown in Table \ref{tab:main_and_ablation}, \textbf{Sensoformer outperforms all ML baselines.} DeepONet struggles significantly (Mean Kagan angle $35.4^\circ$) because mapping extremely sparse, dynamically changing spatial queries into a continuous function space is highly inefficient. MPNN performs better ($27.0^\circ$) but falls short of Sensoformer because seismic radiation patterns are global phenomena; local message-passing over-smooths features and fails to explicitly capture long-range relative phase constraints. 

Furthermore, Sensoformer matches or exceeds recent domain-specific deep learning SOTA. It substantially outperforms previous baselines \citep{ross2018p, cheng2023refined} (typically $>30^\circ$ error). While recent dense-network solvers like FOCONET \citep{song2025foconet} and DiTing \citep{zhao2023ditingmotion} report $20^\circ$--$30^\circ$ errors on highly curated subsets, Sensoformer achieves a Median Kagan angle of $19.7^\circ$ on sparse, ad-hoc operational data. This error effectively hits the label noise floor observed between different human-analyst catalogs \citep{yang2012computing}.

\begin{table}[h]
\caption{\textbf{Main Results and Ablation Studies on Real-World Catalog.} Sensoformer significantly outperforms neural operators, graph-based, and pooling baselines under extreme spatial sparsity. The ablation results further demonstrate that both the architectural decoupling (Dual Towers) and all four rigorous physical augmentations (PSDR) are essential for bridging the sim-to-real gap.}
\label{tab:main_and_ablation}
\begin{center}
\begin{small} 
\begin{tabular}{@{} l c @{\hspace{0.3in}} l c @{}}
\toprule
\textbf{Model / Architecture Variant} & \textbf{Kagan Angle} & \textbf{PSDR Component Ablation} & \textbf{Kagan Angle} \\
& \textit{(Mean / Median)} & & \textit{(Mean / Median)} \\
\midrule
DeepONet (Neural Operator)  & 35.4$^\circ$ / 28.3$^\circ$ & w/o Natural Noise Injection & 29.7$^\circ$ / 22.1$^\circ$ \\
DeepSets (Global Pooling)   & 28.4$^\circ$ / 24.5$^\circ$ & w/o Waveform Distortion ($\mathcal{T}$) & 28.3$^\circ$ / 21.3$^\circ$ \\
MPNN (KNN Graph)            & 27.0$^\circ$ / 22.0$^\circ$ & w/o Var-Geometry Masking ($\mathcal{M}$) & 27.7$^\circ$ / 20.4$^\circ$ \\
Single Tower (w/o P/S Decoupling) & 36.0$^\circ$ / 31.3$^\circ$ & w/o Earth Structure ($\phi$) Rand. & 27.4$^\circ$ / 20.0$^\circ$ \\
\midrule
\textbf{Sensoformer (Ours - Full)} & \textbf{23.9$^\circ$ / 19.7$^\circ$} & \textbf{Sensoformer (Ours - Full)} & \textbf{23.9$^\circ$ / 19.7$^\circ$} \\
\bottomrule
\end{tabular}
\end{small}
\end{center}
\vskip -0.1in
\end{table}

\subsection{Sim-to-Real Generalization}
\begin{figure}[t!]
    \centering
    \includegraphics[trim={0.2cm 0.2cm 0.2cm 0.2cm}, clip, width=0.8\linewidth]{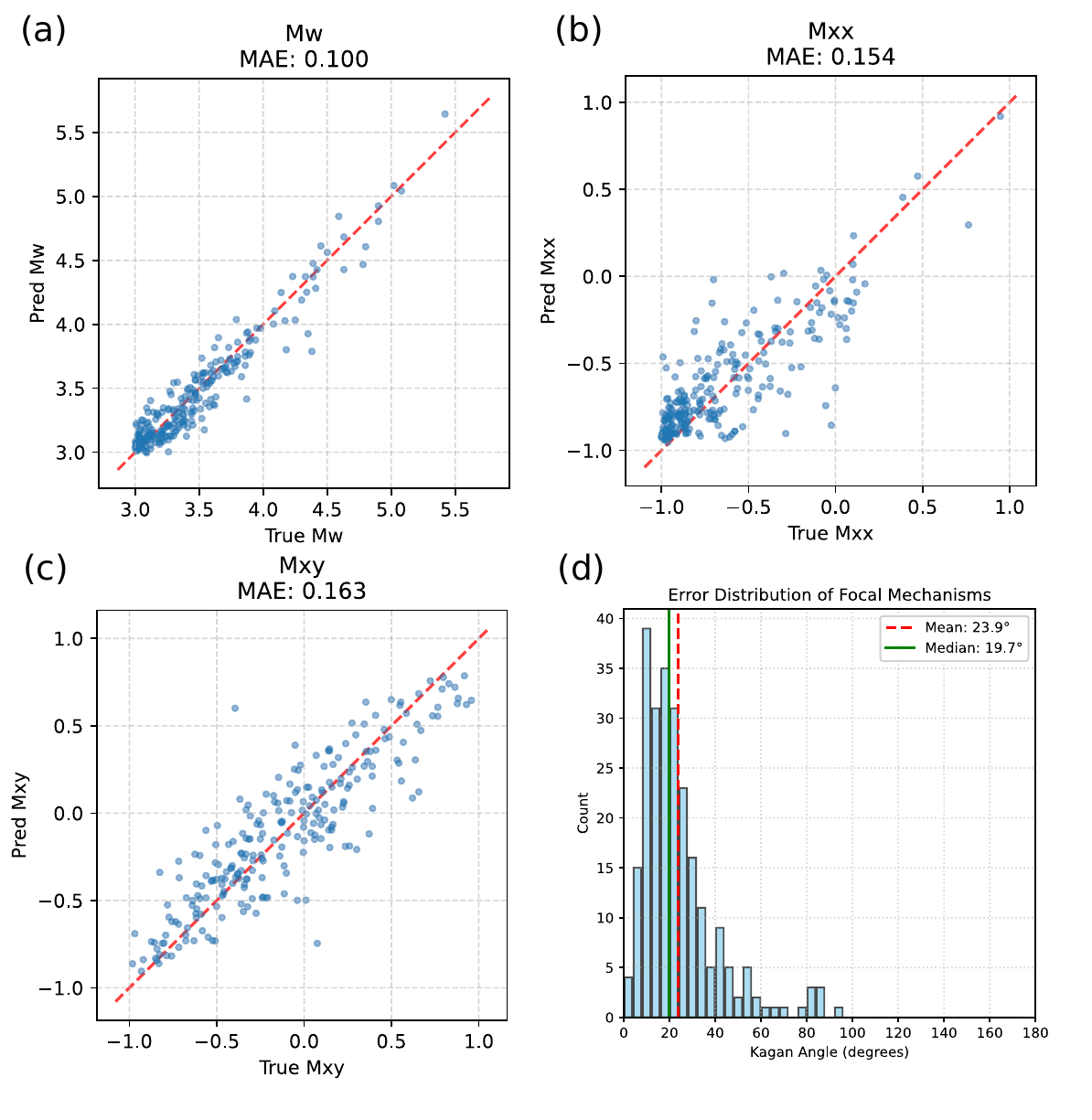} 
    
    \vspace{-0.1in}
    
    \caption{\textbf{Sim-to-Real Inference Performance.} Evaluation on the held-out real-world catalog. \textbf{(a)} Predicted vs. true Magnitude ($M_W$). \textbf{(b-c)} Normalized Moment Tensor components ($M_{xx}, M_{xy}$) maintain robust linearity. \textbf{(d)} The Kagan angle error distribution confirms Sensoformer hits the label uncertainty floor of manual catalogs.}
    \label{fig:performance}
    
    \vspace{-0.25in}
\end{figure}

When applied to real-world data (Figure \ref{fig:performance}), Sensoformer maintains robust linearity despite the profound domain gap. Notably, the magnitude prediction remains highly accurate (MAE=$0.10$) even though real-world waveforms are severely distorted by scattering. This robustness directly validates our multi-modal design, which explicitly disentangles geometric spreading (via the scalar tower) from source energy physics.

\subsection{Ablation Studies}
\label{sec:ablation}
To isolate the contributions of our architectural design and the PSDR pipeline, we performed extensive ablation studies on the real-world dataset (Table \ref{tab:main_and_ablation}). 

\textbf{Architectural Ablations:} We found that compressing P-wave and S-wave streams into a \textit{Single Tower} causes a performance collapse ($36.0^\circ$). This confirms that explicitly disentangling compressional and shear energy is physically necessary to prevent feature confounding. Furthermore, removing the \textit{Scalar Tower} causes the Magnitude ($M_W$) prediction MAE to increase by over 150\% (Appendix D), demonstrating that waveform shapes alone are scale-ambiguous without explicit geometric context.

\textbf{PSDR Ablations:} We systematically stripped each of the four randomization modules from the synthetic pre-training phase. The results indicate that removing any component degrades sim-to-real transfer. Replacing real-world noise with simple white noise (\textit{w/o Natural Noise}) and removing scattering effects (\textit{w/o Waveform Distortion}) caused the largest error increases, highlighting that realistic frequency-dependent distortions are the primary bottleneck for deploying physics-based simulations in the real world.

\section{Interpretability: Attention as a Dynamic Information Bottleneck}
Deep learning models for physical systems are often criticized as black boxes \citep{rudin2019stop}. Here, we show that beyond achieving high empirical accuracy, Sensoformer exhibits highly interpretable internal representations that align with established statistical principles \citep{iten2020discovering}.

\begin{figure}[t!]
    \centering
    \includegraphics[width=0.85\linewidth]{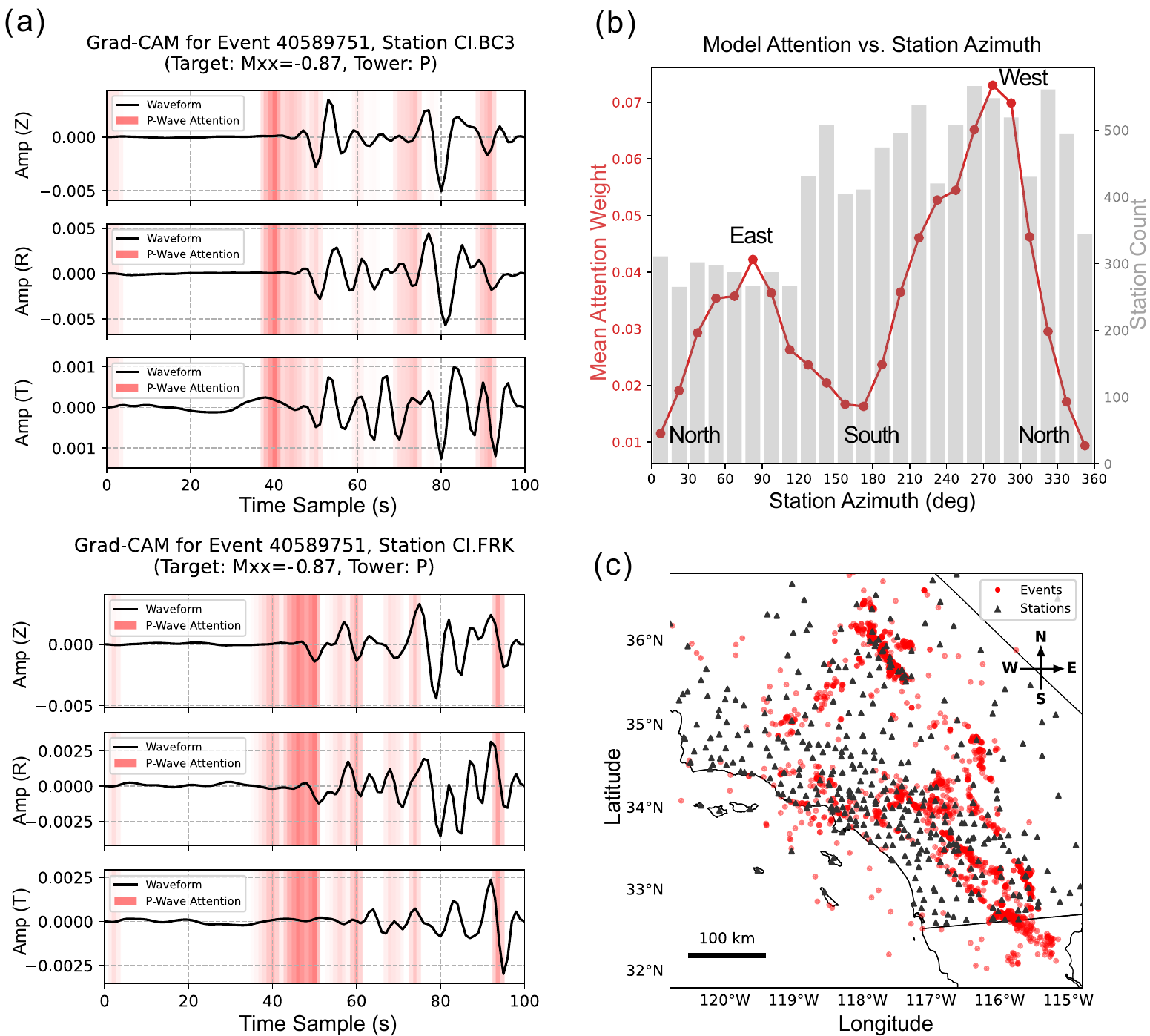} 
    \vspace{-0.15in}
    \caption{\textbf{Interpretability and Sensor Selection.} \textbf{(a)} Local Temporal Attention: Grad-CAM visualization reveals that the model effectively leverages the P-coda and early scattered phases (red regions), extracting constraints from data traditionally discarded as noise. \textbf{(b)} Global Spatial Attention: Aggregated attention weights reveal a learned anisotropy, prioritizing East-West stations. \textbf{(c)} As shown in the spatial map, this learned policy directly counters the network's geometric bias (dense N-S coverage), demonstrating that the model autonomously identifies the spatial information bottleneck.}
    \label{fig:xai}
    \vspace{-0.2in}
\end{figure}

Classical domain solvers typically rely on isolated data points (e.g., the first P-wave arrival), discarding the subsequent complex scattering (coda) as intractable noise. However, our Grad-CAM analysis (Figure \ref{fig:xai}a) reveals that Sensoformer's attention extends substantially into these early scattered phases. From an ML perspective, the model has learned that these transient dynamics contain rich, structured constraints regarding the source mechanism \citep{aki2002quantitative}, effectively leveraging the full temporal context that manual heuristics ignore.

The most compelling behavior emerges in the spatial self-attention mechanism. When aggregating attention scores by station azimuth (Figure \ref{fig:xai}b), we observe a systematic anisotropy: the model consistently upweights stations in the East-West direction ($90^{\circ}/270^{\circ}$) and downweights the highly dense stations to the North and South. Because the pre-training synthetics featured uniformly distributed source mechanisms, this attention bias is strictly driven by the network topology itself.

This spatial weighting demonstrates the spontaneous emergence of an information bottleneck \citep{tishby2000information}. While the N-S sensors are numerous (Figure \ref{fig:xai}c), their spatial correlation results in redundant constraints with diminishing returns. Sensoformer's attention mechanism acts as a dynamic spatial preconditioner: it downweights redundant observations to mitigate multicollinearity while prioritizing under-sampled orthogonal views (E-W) to stabilize the inference. In statistical terms, the architecture implicitly mimics D-Optimal experimental design \citep{pukelsheim2006optimal}, maximizing information gain by focusing on the most variance-reducing elements in the set.

\section{Methodological Insights}

\textbf{Amortized Inference over Ad-Hoc Sets.} Classical physical inversion treats every event as an isolated $O(K)$ optimization problem, which is computationally expensive and highly sensitive to initialization. Sensoformer shifts this paradigm to amortized inference \citep{gershman2014amortized}. By transferring the heavy computational burden to the PSDR pre-training phase, the model learns a global, robust inverse operator allowing $O(1)$ real-time prediction \citep{ongie2020deep}. 

\textbf{Attention as a Relational Inductive Bias.} The failure of pooling (DeepSets) and the suboptimal performance of local message passing (MPNN) underscore the importance of relational inductive biases. Physical wavefields are globally correlated. Rather than relying on rigid graphs or localized neighbors, the all-to-all self-attention mechanism computes a dense similarity matrix \citep{tsai2019transformer}, dynamically evaluating the geometric leverage and phase differences between any two sensors to reconstruct the global radiation pattern.

\textbf{PSDR as Invariant Representation Learning.} From a causal representation perspective, our Physics-Structured Domain Randomization operates as a controlled intervention on nuisance variables \citep{pearl2009causal}. By randomizing the propagation environment (velocity models, noise, topology) while keeping the physical targets fixed, we mathematically regularize the encoder to discard spurious correlations (e.g., local site effects) and learn the underlying invariant physics \citep{arjovsky2019invariant}.

\section{Conclusion}

In this work, we proposed Sensoformer, a robust set-attention framework designed for sim-to-real inference on irregular, variable-geometry sensor arrays. By integrating a multi-modal set architecture with Physics-Structured Domain Randomization, we demonstrated that deep learning can overcome profound distribution shifts and spatial sparsity without relying on massive labeled real-world datasets. Our benchmarking revealed that global self-attention outperforms pooling, MPNN, and Neural Operator baselines on unstructured physical data. Furthermore, interpretability analysis showed that the model dynamically identifies spatial information bottlenecks, mimicking optimal sensor selection. By providing a scalable approach to reason over ad-hoc observations, Sensoformer offers a generalizable blueprint for deploying reliable AI across the physical sciences and industrial IoT.

\section*{Limitations and Broader Impact}

\textbf{Limitations.} While PSDR significantly bridges the sim-to-real gap, the model assumes that the real-world physics remains within the support of the randomized simulation library. Out-of-distribution (OOD) scenarios, such as geological regions with extreme 3D heterogeneities unrepresented in the pre-training distributions, may lead to performance degradation. Additionally, the current framework provides deterministic point estimates; extending the architecture to output calibrated predictive distributions (e.g., via Bayesian Neural Networks or conformal prediction) for robust uncertainty quantification remains an important direction for future work. 

\textbf{Broader Impact.} Sensoformer advances the deployment of robust machine learning in physical monitoring systems. In our benchmarking domain, it provides a highly scalable solution that can directly accelerate earthquake early warning and hazard mitigation. There are no anticipated negative societal impacts or dual-use ethical risks associated with this work, as the methodology and application are strictly focused on environmental and geophysical monitoring.

\begin{ack}
\end{ack}

\bibliographystyle{plainnat}
\bibliography{example_paper}

\newpage
\appendix
\section{Implementation Details and Hyperparameters}
\label{app:implementation}

To ensure full reproducibility, we provide the specific architectural specifications, training hyperparameters, and the exact configurations for our Physics-Structured Domain Randomization (PSDR) framework. The framework was implemented in PyTorch using a modular design to support dynamic configuration management. All experiments were executed on NVIDIA A100 (40GB) GPUs.

\subsection{Physics-Structured Domain Randomization (PSDR) Configurations}
\label{app:psdr_details}
As introduced in the main text, bridging the sim-to-real gap relies heavily on explicitly randomizing the underlying physical generative processes. The specific hyperparameters for the four PSDR modules are defined as follows:

\begin{itemize}
    \item \textbf{Generative Environment ($\phi$):} We extracted 17 distinct 1D velocity profiles at a $1^{\circ}$ spatial resolution from the CRUST1.0 database spanning the Southern California region (latitudes $33.0^{\circ}\text{N}$ to $37.0^{\circ}\text{N}$, longitudes $121.0^{\circ}\text{W}$ to $115.0^{\circ}\text{W}$). During synthetic generation, one profile is uniformly sampled, $v \sim \mathcal{U}(1, 17)$, to compute the Green's functions.
    \item \textbf{Signal Distortion ($\mathcal{T}$):} We apply rigorous distortions to the analytical synthetics:
    \begin{itemize}
        \item \textit{Travel-time Shifts:} The initial P-wave arrival time is randomly shifted by $\Delta_{t} \sim \mathcal{U}(-8\%, +8\%)$.
        \item \textit{Amplitude Scaling:} P-wave and S-wave amplitudes are independently scaled by factors of $\alpha_P \sim \mathcal{U}(0.5, 2.0)$ and $\alpha_S \sim \mathcal{U}(0.5, 3.0)$.
        \item \textit{Synthetic Coda Injection:} We dynamically generate and inject an exponentially decaying coda. We generate truncated Gaussian white noise, filter it using a 4th-order Butterworth bandpass filter ($0.1$--$2.0\text{ Hz}$), and modulate it with an envelope $E(t) = A \exp(-bt)$. The initial coda amplitude $A \sim \mathcal{U}(0.5, 1.0) \times \max(|S_{\text{wave}}|)$, and the decay rate is $b \sim \mathcal{U}(0.1, 0.3)$.
    \end{itemize}
    \item \textbf{Realistic Noise Injection ($\mathbf{n}$):} We compiled a noise library containing over 50,000 real seismic noise traces extracted from pre-event windows of actual Southern California recordings. A randomly sampled trace is superimposed onto the synthetic waveform, scaled such that its maximum amplitude does not exceed $0.3 \times$ the maximum amplitude of the clean synthetic signal.
    \item \textbf{Variable-Geometry Masking ($\mathcal{M}$):} For any synthetic event with more than 50 available stations, we randomly discard stations to retain an integer number $N \sim \mathcal{U}(30, 50)$, enforcing robust representation learning under dynamic spatial sparsity.
\end{itemize}

\subsection{Model Architecture and Training Protocol}
Sensoformer adopts a hybrid architecture combining CNNs for local waveform feature extraction and Set Transformers for global spatial aggregation. The model consists of approximately 1.5M trainable parameters. 

\begin{table}[h]
\centering
\caption{\textbf{Sensoformer Hyperparameters.} Values are chosen based on grid search validation performance.}
\label{tab:hyperparams}
\begin{tabular}{@{}ll@{}}
\toprule
\textbf{Parameter} & \textbf{Value} \\ \midrule
\multicolumn{2}{c}{\textit{Multi-Modal Station Encoder (Siamese 1D-ResNet)}} \\
Input Channels & 6 (P-wave) + 6 (S-wave) \\
Scalar Feature Dim & 20 (Continuous Spatial Encoding) \\
Conv Kernel Size & 7 \\
ResNet Blocks & 3 \\
Station Embedding Dim ($d_{model}$) & 128 \\ \midrule
\multicolumn{2}{c}{\textit{Event Aggregator (Set Transformer)}} \\
Layers & 3 \\
Attention Heads & 4 \\
Feedforward Dimension & 256 \\
Dropout Rate & 0.1 \\
Positional Encoding & None (Permutation Invariant) \\ \midrule
\multicolumn{2}{c}{\textit{Stage 1: Synthetic Pre-training}} \\
Dataset Size & 100,000 synthetic events \\
Batch Size (Dynamic Padding) & 512 \\
Optimizer & AdamW \citep{loshchilov2017decoupled} \\
Learning Rate & $2 \times 10^{-4}$ \\
Loss Function & MSE Loss \\
Early Stopping Patience & 30 epochs \\ \midrule
\multicolumn{2}{c}{\textit{Stage 2: Real-world Fine-tuning}} \\
Dataset Size & $\sim$1,700 training events \\
Batch Size & 64 \\
Optimizer & AdamW \\
Learning Rate (Backbone \& Heads) & $2 \times 10^{-6}$ \\
Loss Function & Focal L1 Loss ($\gamma=1.5, \beta=1.0$) \\
Sampling Strategy & WeightedRandomSampler \\ 
Early Stopping Patience & 50 epochs \\ \bottomrule
\end{tabular}
\end{table}

\textbf{Dynamic Batching:} Unlike standard fixed-size inputs, Sensoformer handles a variable number of sensors per event. We implement a dynamic collate function that pads batches to the maximum sensor count within the current batch (which is subsequently masked during attention computation), effectively optimizing GPU memory usage without distorting the set distribution.

\clearpage

\section{Pre-training Dynamics and Synthetic Benchmark Evaluation}
\label{app:synthetic_results}

We evaluate Sensoformer on a held-out test set of synthetic events generated using the PSDR engine but featuring unseen source parameters and geometries.

\begin{figure}[h!]
    \centering
    \includegraphics[width=0.9\linewidth]{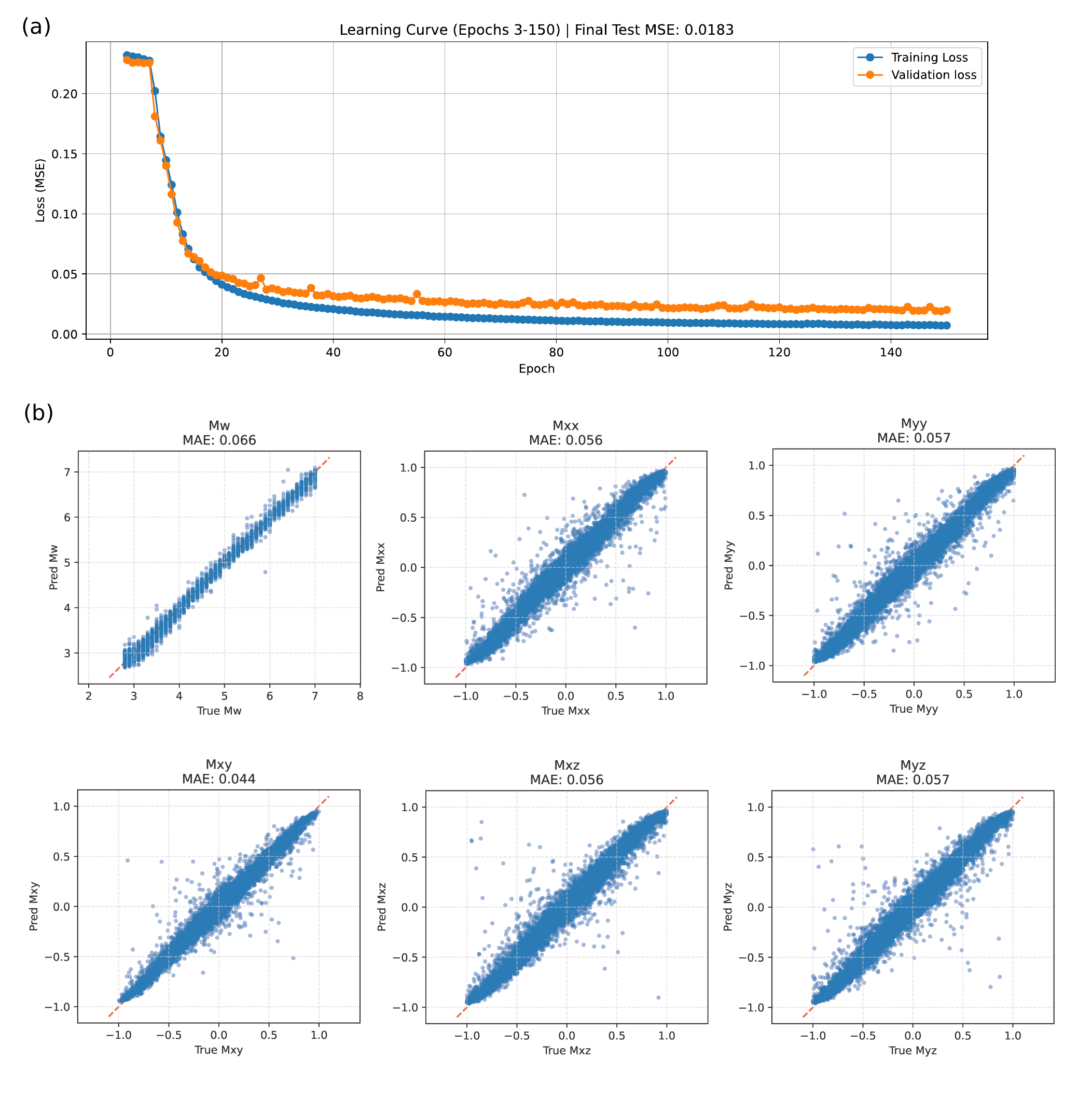}
    \caption{\textbf{Synthetic Pre-training Performance.} \textbf{(a)} Training and validation loss curves (MSE) over 150 epochs. \textbf{(b)} Parity plots for Magnitude ($M_w$) and Moment Tensor components ($M_{xx}, M_{yy}, \dots$) on the synthetic test set. The tight alignment along the $y=x$ diagonal confirms the model's capacity to invert physics-based simulations.}
    \label{fig:pretrain_results}
\end{figure}

Figure \ref{fig:pretrain_results} illustrates the training dynamics and regression performance on the synthetic domain. The learning curve (Figure \ref{fig:pretrain_results}a) shows smooth convergence without overfitting. The parity plots (Figure \ref{fig:pretrain_results}b) reveal near-perfect linearity for both Moment Magnitude ($M_w$) and Moment Tensor components ($M_{ij}$). The Mean Absolute Error (MAE) for magnitude is extremely low (0.066), confirming that the continuous spatial encoding correctly disentangles geometric spreading from source energy.

\begin{figure}[h!]
    \centering
    \includegraphics[width=0.6\linewidth]{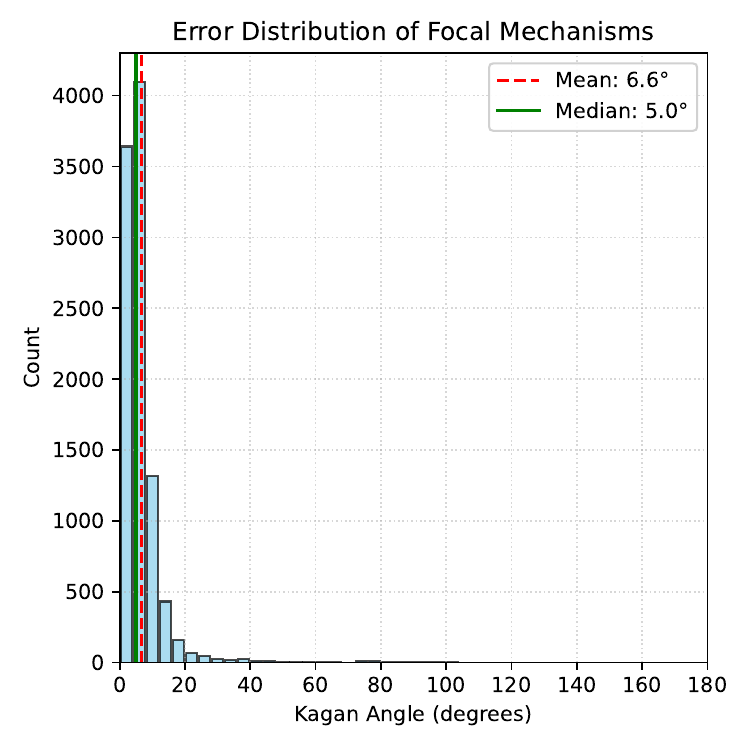}
    \caption{\textbf{Error Distribution on Synthetic Data.} Histogram of Kagan Angle errors. The extremely low mean error ($6.6^\circ$) and median ($5.0^\circ$) demonstrate the high fidelity of the set-attention architecture in the absence of severe domain shift.}
    \label{fig:pretrain_kagan}
\end{figure}

Figure \ref{fig:pretrain_kagan} presents the error distribution of the physical predictions. The model achieves a mean Kagan angle of $6.6^\circ$ and a median of $5.0^\circ$. This implies that in a controlled physical environment, Sensoformer can recover the complex spatial mechanism with high fidelity. The performance gap between this synthetic domain ($6.6^\circ$) and the real-world domain ($23.9^\circ$, Main Text) accurately quantifies the residual ``Reality Gap'' caused by extreme, unmodeled 3D heterogeneities.

\begin{figure}[h!]
    \centering
    \includegraphics[width=0.95\linewidth]{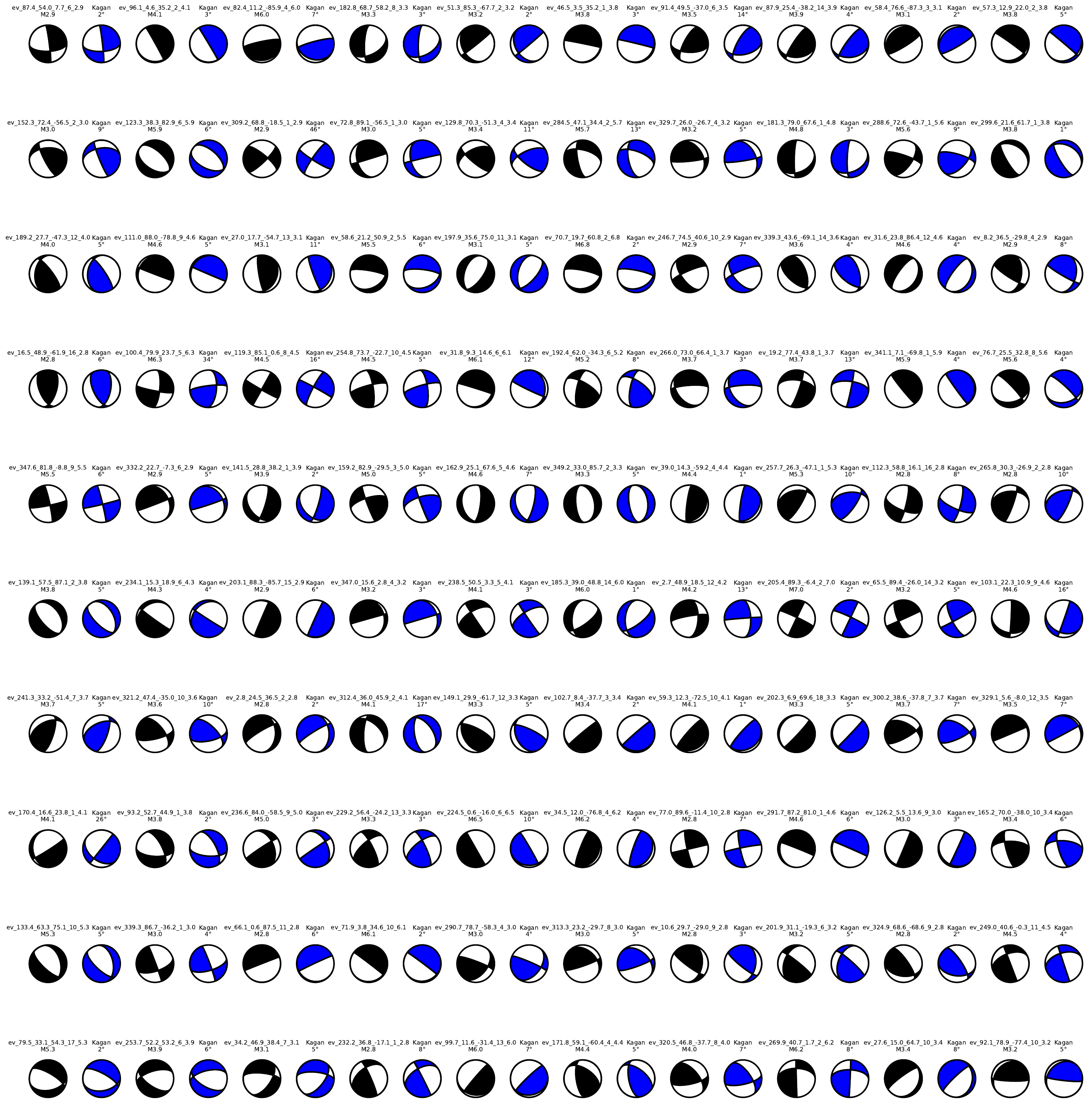}
    \caption{\textbf{Visualizing Focal Mechanisms (Synthetic).} A grid comparison of ground truth (black) vs. Sensoformer predicted (blue) geometric mechanisms for a random subset of synthetic test events.}
    \label{fig:pretrain_beachballs}
\end{figure}

Figure \ref{fig:pretrain_beachballs} provides a visual comparison of the predicted vs. ground truth spatial mechanisms (commonly visualized as "beachballs" in geophysics) for held-out events. The visual agreement is consistent across various physical faulting types, confirming that the network successfully avoids mode collapse.

\clearpage

\newpage

\section{Real-World Fine-Tuning and Sim-to-Real Transfer Dynamics}
\label{app:finetune_results}

Following PSDR pre-training, we fine-tune Sensoformer on a small set of real-world events ($\sim$1,700 training samples) from the target distribution. 

\begin{figure}[h!]
    \centering
    \includegraphics[width=0.9\linewidth]{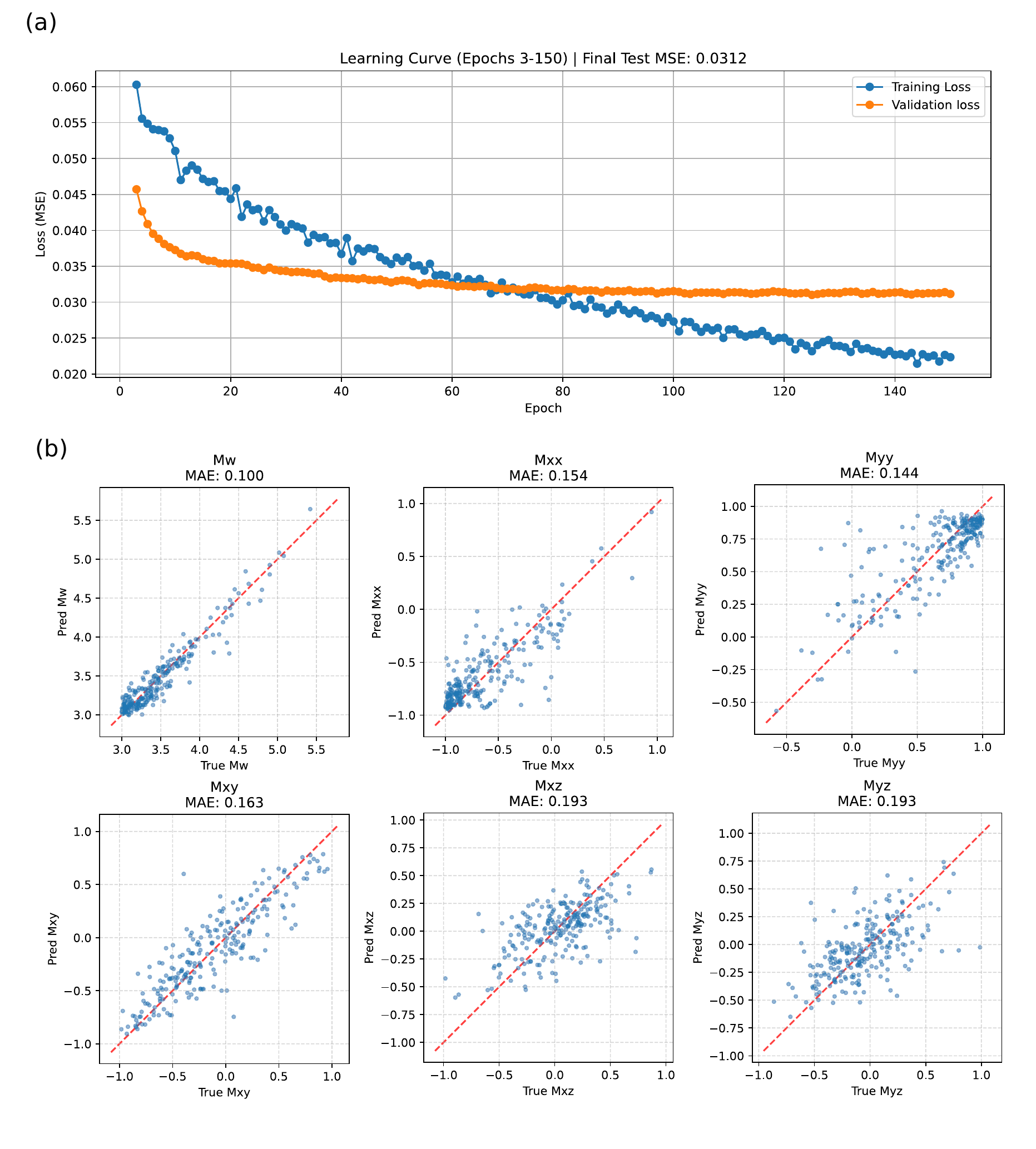}
    \caption{\textbf{Real-World Fine-tuning Performance.} \textbf{(a)} Learning curves during the fine-tuning stage. \textbf{(b)} Parity plots on the held-out real-world test set. Magnitude ($M_w$) prediction remains highly robust (MAE$=0.11$) despite significant waveform distortions.}
    \label{fig:finetune_results}
\end{figure}

Figure \ref{fig:finetune_results}a shows the learning curve during fine-tuning. Despite the small dataset size, the validation loss decreases steadily and stabilizes, confirming that the invariant physical features learned during PSDR pre-training act as a strong regularizer against overfitting. Figure \ref{fig:finetune_results}b shows the parity plots for real data. While there is increased variance compared to the synthetic domain (attributable to real-world noise and manual label uncertainty), the predictions remain robustly unbiased and linear.

\begin{figure}[h!]
    \centering
    \includegraphics[width=0.95\linewidth]{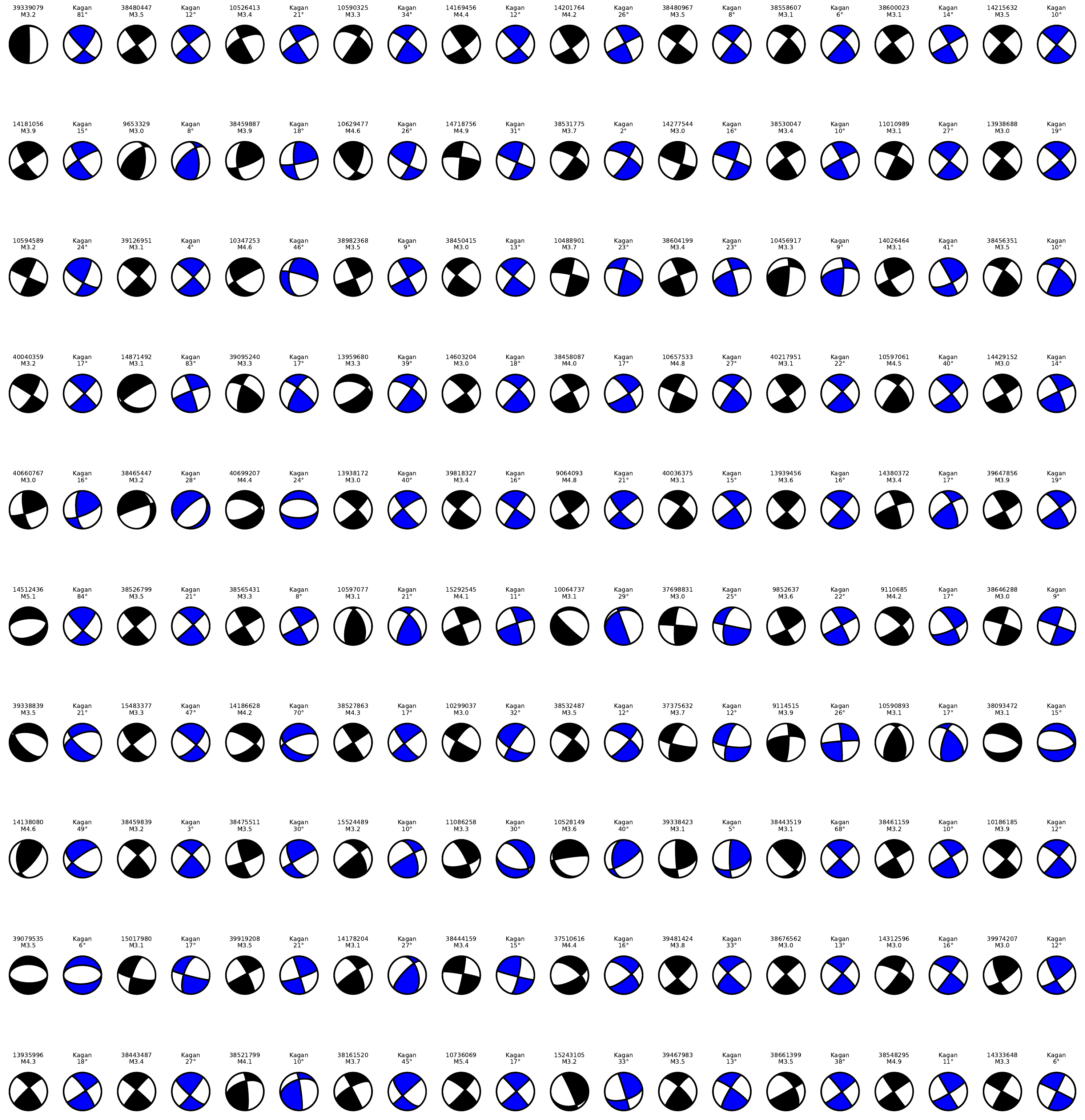}
    \caption{\textbf{Visualizing Focal Mechanisms (Real World).} Comparison of manual catalog solutions (Black) vs. Sensoformer predictions (Blue) for real-world events. The model successfully recovers complex geometries even under sparse coverage and high noise.}
    \label{fig:finetune_beachballs}
\end{figure}

Figure \ref{fig:finetune_beachballs} provides a qualitative comparison for held-out real-world events. Sensoformer produces physical solutions that closely match the high-quality manual catalog, demonstrating robust sim-to-real transfer under extreme spatial sparsity.
\clearpage

\newpage

\section{Extended Ablation Visualizations}
\label{app:ablation_viz}

In Section 5 of the main text, we summarized the impact of removing key architectural components via quantitative metrics. Here, we provide detailed visualizations to qualitatively understand the specific failure modes of these ablated model variants.

\subsection{Impact of Continuous Spatial Encoding (Multi-Modal Fusion)}
\begin{figure}[h!]
    \centering
    \includegraphics[width=0.9\linewidth]{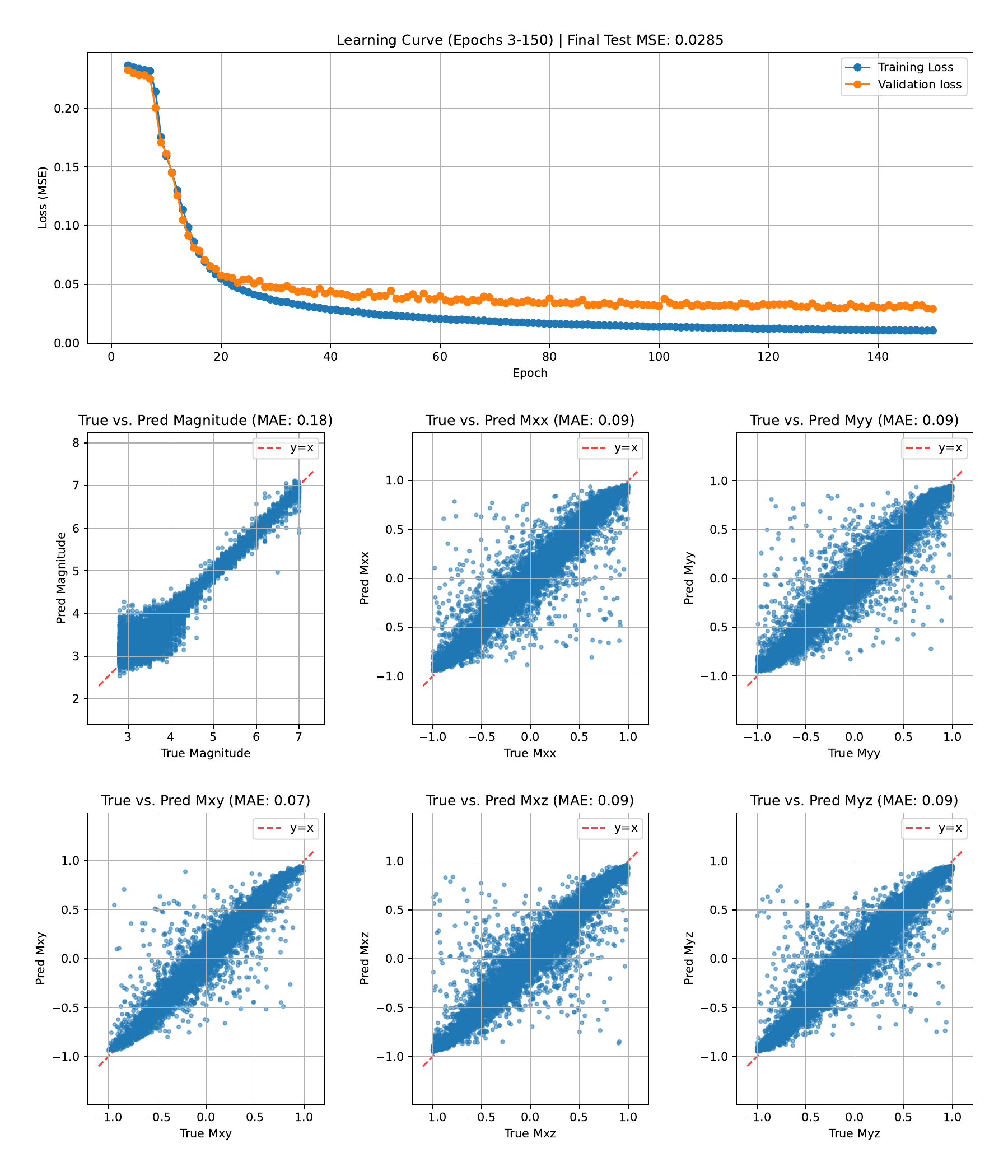}
    \caption{\textbf{Ablation: w/o Continuous Spatial Encoding.} Parity plots for the model trained without scalar metadata. Note the degradation in Magnitude ($M_w$) prediction (MAE $0.07 \to 0.18$), visually confirming the necessity of spatial feature fusion.}
    \label{fig:ablation_no_scalar}
\end{figure}

We trained a variant of Sensoformer without the Scalar Tower (i.e., using only raw waveforms as input). As shown in Figure \ref{fig:ablation_no_scalar}, removing the continuous spatial encoding causes a catastrophic collapse in energy scale prediction. The Magnitude ($M_w$) scatter plot (top left) exhibits high variance and poor correlation compared to the full model. This visually confirms our hypothesis: waveform shapes alone are scale-ambiguous due to the trade-off between physical source energy and spatial distance spreading. Explicit multi-modal fusion of spatial metadata is absolutely essential.
\clearpage

\subsection{Impact of Self-Attention (Relational Reasoning)}
\begin{figure}[h!]
    \centering
    \includegraphics[width=0.9\linewidth]{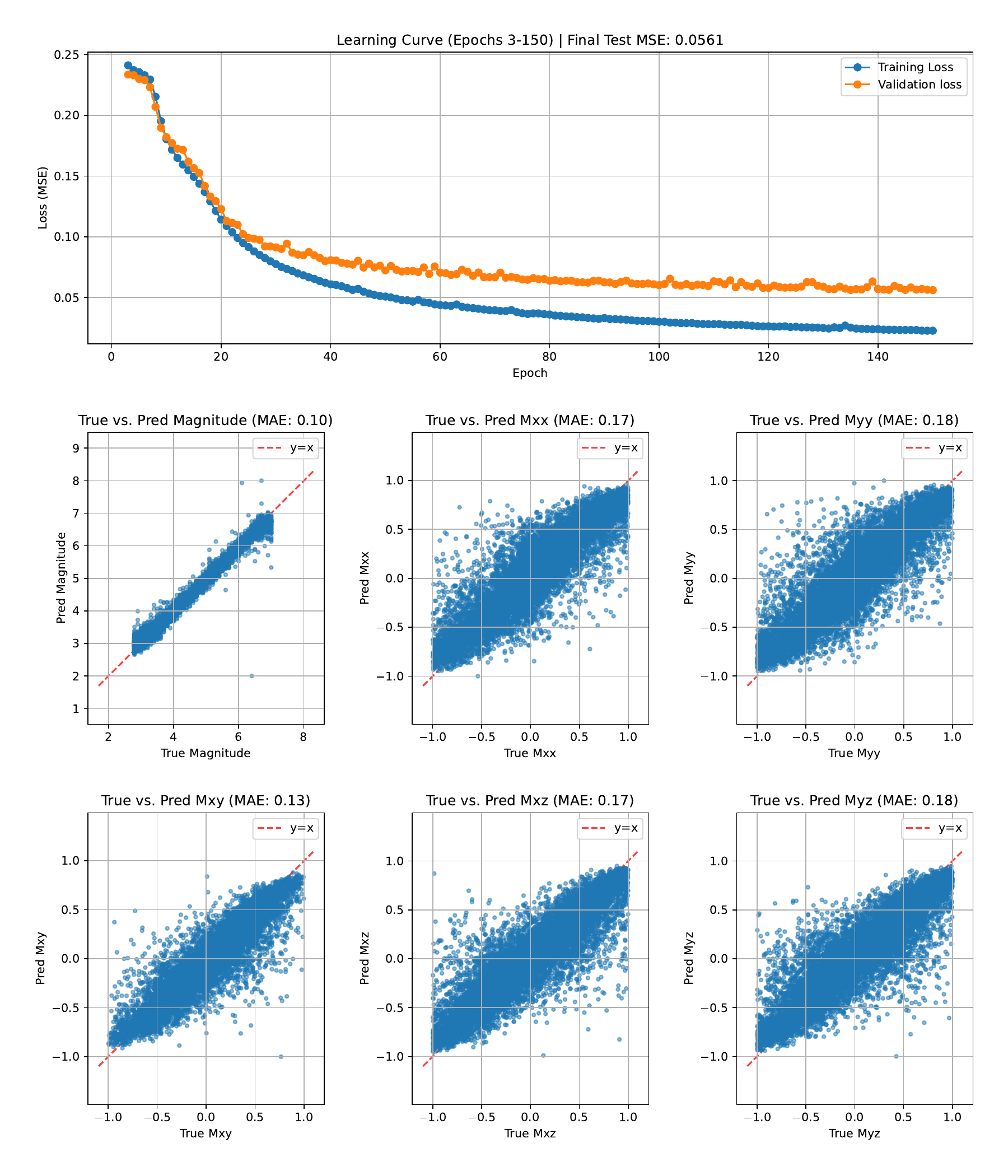}
    \caption{\textbf{Ablation: w/o Self-Attention (DeepSets).} Parity plots for the model using simple global pooling instead of all-to-all attention. The estimation of physical components degrades significantly (Mean Tensor MAE $0.06 \to 0.17$), illustrating the failure to capture global relational physics.}
    \label{fig:ablation_deepsets}
\end{figure}

We trained a variant where the Set Transformer was replaced by a DeepSets-style architecture (independent processing followed by global mean pooling). As shown in Figure \ref{fig:ablation_deepsets}, the physical tensor components ($M_{xx}, M_{yy}, \dots$) exhibit higher scatter and bias compared to the full Sensoformer model. Pooling-based architectures independently aggregate features, fundamentally failing to capture the pairwise relative phase information (e.g., phase polarity reversals between distant sensors) required to distinguish complex spatial geometries. Global self-attention provides the necessary relational inductive bias to resolve these spatial ambiguities.

\end{document}